# A New Chinese Landscape Paintings Generation Model based on Stable Diffusion using DreamBooth

[a]Yujia Gu, California State University, Long Beach, United States
[b]Xinyu Fang, Columbia University, New York, United States
[c]Xueyuan Deng, Northeastern University, Seattle, United States
[d]Zihan Peng, No.1 Central Primary School Shanghai Huangpu, Shanghai, China
[*] Yinan Peng, Shanghai Palmim Information Technology Ltd, Shanghai, China
Yujia Gu and Xinyu Fang are first co-authors, corresponding email: kevin@palmim.com

This study mainly introduces a method combining the Stable Diffusion Model (SDM) and Parameter-Efficient Fine-Tuning method for generating Chinese Landscape Paintings. This training process is accelerated by combining LoRA with pre-trained SDM and DreamBooth with pre-trained SDM, respectively. On the Chinese Landscape Paintings Internet dataset used in this paper, this study finds that SDM combined with DreamBooth exhibits superior performance, outperforming other models, including the generic pre-trained SDM and LoRA-based fine-tuning SDM. The SDM combined with DreamBooth achieves a FID of 12.75 on the dataset and outperforms all other models in terms of expert evaluation, highlighting the model's versatility in the field of Chinese Landscape Paintings given the unique identifier, high fidelity and high quality. This study illustrates the potential of specialised fine-tuning method to improve the performance of SDM on domain-specific tasks, particularly in the domain of Landscape Paintings.

**Keywords**: Chinese Landscape Paintings, Stable Diffusion Model, DreamBooth Fine-tuning, LoRA Fine-Tuning

## 1.INTRODUCTION

Chinese landscape painting image generation by Text-to-Image models has seen significant advancements in recent years. Various studies have explored different approaches to improve the generation of Chinese landscape paintings using text prompts. One common challenge faced by researchers is the need for a large number of high-quality text-image pairs for training these models [1]. To address this issue, some studies have proposed methods that do not require text data for training text-to-image generation models [1]. This approach opens up new possibilities for generating Chinese landscape paintings without relying on extensive text datasets. In the realm of text-to-image generation, the use of fine-tuning methods has been crucial in enhancing the performance of models. For instance, Hao et al. introduced prompt adaptation, a framework that automatically adjusts user input to align with model-preferred prompts, leading to improved generation results [2]. Additionally, Wang et al. proposed CCLAP, a controllable Chinese landscape painting generation method that leverages Latent Diffusion Models for generating paintings with specific content and style [3]. These parameter-efficient fine-tuning techniques have proven to be effective in enhancing the quality and controllability of Chinese landscape painting generation models. Moreover, the integration of meta-learning techniques has shown promise in improving the generalization capability of text-to-image models. Chen et al. introduced MLFont, a deep meta-learning-based font generation method that leverages existing fonts to enhance the model's ability to generate new fonts with few-shot learning [4]. This approach demonstrates the potential of meta-learning in addressing the challenges of generating diverse Chinese landscape paintings using Text-to-Image models. Furthermore, the concept of explainable image quality evaluation has gained traction in the evaluation of text-to-image generation methods. Chen et al. proposed X-IQE, a novel approach that utilizes visual large language models to evaluate text-to-image generation methods by generating textual explanations [5]. This method offers advantages such as distinguishing between real and generated images, evaluating text-image alignment, and assessing image aesthetics without the need for model training or fine-tuning. In addition to these advancements, studies have also focused on enhancing the personalization and stylization capabilities of text-to-image models. Li et al. introduced block-wise Low-Rank Adaptation (LoRA) to perform fine-grained fine-tuning for different blocks of SD, enabling effective personalization and stylization in text-to-image generation [6]. This approach aims to generate images that are faithful to input prompts and target identity while incorporating desired styles, showcasing the potential for creating highly personalized Chinese landscape paintings. Overall, the literature on Chinese landscape painting image generation by Text-to-Image models highlights the importance of parameter-efficient fine-tuning methods, meta-learning techniques, and explainable image quality evaluation in advancing the field. These approaches have contributed to the development of more controllable, personalized, and high-quality Chinese landscape painting generation models, paving the way for further innovations in this domain. In light of the preceding research, this paper will concentrate on identifying the optimal method for efficiently fine-tuning the Diffusion Model for generating the gorgeous Chinese landscape paintings.

## 2.RELATED WORK

**Stable Diffusion Model:** The original diffusion model was introduced by Song and Ermon [7], who proposed a framework that could be trained to reverse the process of adding Gaussian noise to an image. This approach was further refined by other researchers, leading to the development of the denoising diffusion probabilistic models (DDPMs). These models have been shown to be effective in generating images that are diverse and visually appealing.

The Stable Diffusion Model builds upon the foundation laid by DDPMs by incorporating additional stability features. This model is designed to mitigate the common issues of mode collapse and image degradation that are often observed in GANs. By leveraging the power of diffusion processes, the Stable Diffusion Model is capable of generating images that are not only diverse but also maintain a high level of fidelity to the original data distribution. Recent advancements in the field have seen the integration of the Stable Diffusion Model with other deep learning techniques, such as variational autoencoders (VAEs) and transformers. This has led to the creation of hybrid models that combine the strengths of both approaches, resulting in even more robust and versatile generative models [8].

**LoRA Micro-Tuning Technique:** In the realm of deep learning, fine-tuning pre-trained models is a common practice to adapt them to specific tasks. However, the full fine-tuning of large models can be computationally expensive and prone to overfitting. To address this, the Low-Rank Adaptation (LoRA) technique has been proposed as an efficient alternative for large language models [9]. LoRA introduces a low-rank structure to the model's weight matrices, allowing for a more targeted and less resource-intensive update of the model parameters. This approach has been shown to be particularly effective in scenarios where only a subset of the model's layers needs to be adapted to the new task.

**DreamBooth Fine-tuning Technique:** The DreamBooth fine-tuning technique is an innovative approach that has emerged in the field of deep learning, offering a way to tailor pre-trained models to perform exceptionally well on specific tasks or datasets [10]. This method is particularly useful when the available training data is limited. The core idea behind DreamBooth is to create a "dream" dataset by training a generative model on a small set of real images, which are then used to generate a large number of augmented images. These synthetic images, when combined with the original data, provide a more diverse and extensive dataset for training the model. This approach has been shown to improve the model's ability to generalize and adapt to new, unseen data [10].

# 3. METHODOLOGY

## 3.1 Latent Diffusion Model

Stable Diffusion Model V1 our paper used is based on the Generative Model. Generative Model includes GAN series, VAE or Flow-based series and Diffusion Model series.

Table 1. DIFFERENCES AMONG GENERATIVE MODELS

| Models | Likelihood-based | Advantages |
|---|---|---|
| GANs | × | Efficient sampling |
| VAE/FLOW-based | √ | Capture data distribution |
| DIFFUSION MODELs | √ | Capture data distribution, High fidelity |

And the Stable Diffusion Model just based on Latent Diffusion, which is a type of Diffusion Models. The principle for Diffusion Model is shown below,

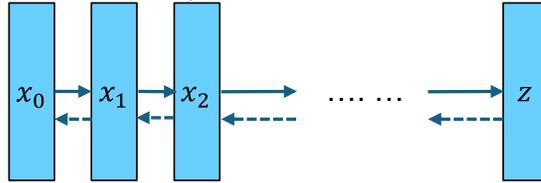

Figure 1. Description for Diffusion Model

As shown in Figure 1, Latent Diffusion will be $x$ through a little bit of noise, into the last close to the distribution of $Z$, which is also called latent representation, with a little bit of noise, this latent representation Finally almost become a pure noise, this is Diffusion Process, and step by step denoising, that is, reverse denoising process.

In more detail, the training of Diffusion Model consists of two processes: Forward Diffusion Process: adding noise to the image. Reverse Diffusion Process: removing noise from the image, this process is also known as denoising and sampling.

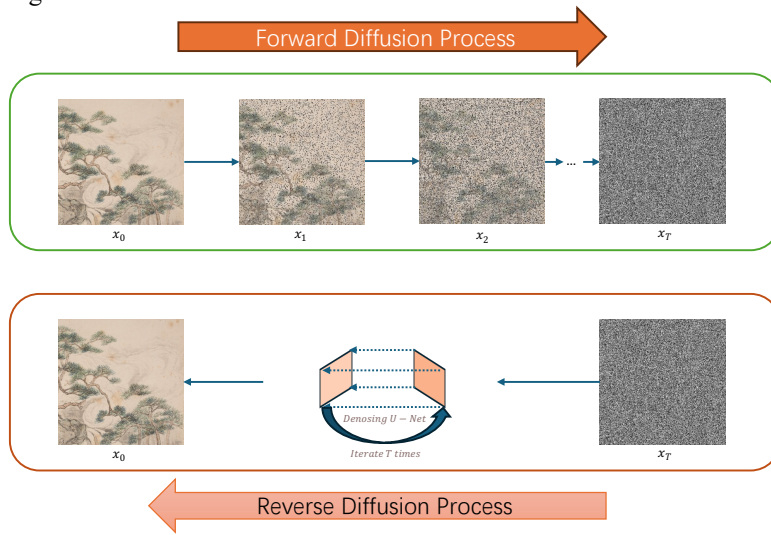

Figure 2. Training Process of Diffusion Model

As shown in Figure 2, the Forward Diffusion Process can be described the following pseudo code:
1. Sample out from $x_0$ sample and add noise.
2. Sample $t$ at time step $T$ and can Embedding it.
3. Generate random noise that follows a normal distribution.
4. Use network prediction noise, which normally use U-Net. And to compute the MSE between the true noise and the estimated noise, which is shown as follows:

$$\nabla_\theta \|\epsilon - \epsilon_\theta(\sqrt{\bar{\alpha}_t}x_0 + \sqrt{1-\bar{\alpha}_t}\epsilon, t)\|^2$$

Where $x_0 \sim q(x_0)$, $t \sim Uniform(\{1,...,T\})$, $\varepsilon \sim N(0,I)$ and $\alpha_t$ is a constant.
5. Backpropagation for gradient updating until convergence.
And then the denoising process as shown:
1. Sample $x_t$ from a Gaussian distribution.
2. Iterate in the order $T,...,1$ in order of iteration.
3. If $t = 1$ make $z = 0$. if $t > 1$, z obeys a Gaussian distribution.
4. Find the mean and variance and hence $x_{t-1}$ using Eq.

$$\mathbf{x}_{t-1} = \frac{1}{\sqrt{\alpha_t}}\left(\mathbf{x}_t - \frac{1-\alpha_t}{\sqrt{1-\bar{\alpha}_t}}\epsilon_\theta(\mathbf{x}_t, t)\right) + \sigma_t \mathbf{z}$$

Where $\epsilon_\theta(\mathbf{x}_t, t)$ always indicate U-Net.
5. Recover $x_0$ after the above iterations.

Then, there is a reformed version of Diffusion Model, which is Latent Diffusion Models. On the other hand, it reduces the consumption of computational resources by applying the diffusion process on the lower dimensional latent space instead of using the actual pixel space.

The Stable Diffusion used in this paper has the same idea as Latent Diffusion, except that Stable Diffusion borrows from Google's Imagen and uses the CLIP ViT-L/14 composition text encoder, which is pre-trained and frozen throughout the training process, and its role is to inject our textual prompt prompt as a condition, injected into the denoising generation process, the other and Latent Diffusion is the same.

Stable Diffusion first trains a powerful pre-trained autoencoder, which learns a potential space that is much smaller than the pixel space, and training the Diffusion Model in this potential space greatly reduces the computational power requirement. The model then introduces a cross-attention layer into the model architecture to enable multimodal training, which can be a more general form of conditional injection, i.e., it can inject text, bounding boxes, and images into the Diffusion Model in a unified way.

### 3.2 Dreambooth Method for Fine-tuning

As shown in Figure 3, the low-resolution text-to-image model was first fine-tuned with input image and text cues containing a unique identifier followed by the subject's class name (e.g., 'a [V] landscape painting'). To prevent overfitting and linguistic drift from causing the model to associate class names (e.g., 'landscape painting') with specific instances, a class-specific prior preservation loss is proposed, which takes advantage of the semantic prior of the classes embedded in the model and encourages it to generate instances of classes with the same unique identifier but with different contexts.

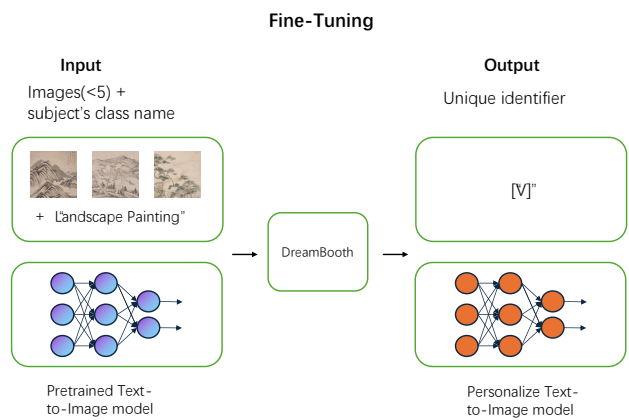

Figure 3. DreamBooth Fine-tuning Method

For example, Chinese landscape paintings dataset will be along with subject's class name by batch, which is "Landscape Painting" here. And then, before those were put into Pretrained Text-to-Image model, it will add unique identifier in front of the subject's class, such as "{{class}}". So the whole prompt is like "{{class}} Landscape Painting". Comparing LoRA and DreamBooth, the advantage of LoRA is that it can fine-tune the pre-trained model on Chinese Landscape Paintings through the low-rank matrix decomposition technique, which is able to improve the adaptability and flexibility of the model while keeping the number of model parameters unchanged, while the advantage of DreamBooth is that it is able to train the model on a small amount of specific The advantage of DreamBooth is that it can train the model on a small amount of specific data (e.g., a dataset such as Chinese Landscape Paintings) to achieve high-quality generation of specific content, which improves the diversity and personalisation of the generated images.

## 4. EXPERIMENT

### 4.1 Dataset and Model Description

There are 8308 images in the dataset used in this paper, with two types of prompts: 'A picture of Chinese Landscape Painting' and 'A picture of Modern Landscape Painting'. The former has 1973 pictures and the latter consists of 6335 pictures. The size of the pictures is 256*256, and they are normalised.

The pre-trained Text-to-Image model used in this paper is 'Stable-Diffusion-v1-5', which checkpoint was initialized with the weights of the Stable-Diffusion-v1-2 checkpoint and subsequently fine-tuned on 595k steps at resolution 512x512 on "laion-aesthetics v2 5+" and 10% dropping of the text-conditioning to improve classifier-free guidance sampling.

### 4.2 Evaluation Metric

Generative models produce high-dimensional and complex structured data, they are different from discriminative models, and it is difficult to evaluate the models with simple metrics. Currently, the more mainstream and reliable metrics for evaluating generative models (discriminative images only) are IS (Inception Score) and FID (Frechet Inception Distance score). The FID score is an improvement on IS. The difference between FID and IS is that IS evaluates the generated image directly, with the larger the value of the metric the better, whereas the FID score generates the evaluation score by comparing the generated image with the real image and calculating a 'distance value', the smaller the value of the metric the better. This paper uses the FID metrics to calculate the average similarity between the generated Chinese Landscape Paintings and real pictures. Meanwhile, to measure the stylistic invariance and background diversity of Chinese Landscape Paintings, this paper also adopts expert manual evaluation.

### 4.3 Experiment Analysis

In the experimental setup, the pre-trained Stable Diffusion model is used to generate an image based on the prompt 'A picture of Chinese Landscape Painting', and then a random picture in the training set is used for FID calculation. This process is repeated 10 times to get the average FID result. Similarly, for DreamBooth and LoRA fine-tuned models, the same strategy is adopted to calculate the FID value, it is worth noting that the DreamBooth fine-tuned model inference needs to be added to the unique identifier, and the training set to maintain consistency.

Table 2: FID RESULTS FOR DIFFERENT MODELS

| Data | Description |
|---|---|
| Pre-trained Stable-Diffusion Model | 26.78 |
| DreamBooth Fine-tuned SDM | 12.75 |
| LoRA Fine-tuned SDM | 17.85 |

In theory, a smaller FID value means that the model outputs an image based on a certain class that is more similar to an image based on this class, and vice versa. From TABLE II, we can see that the fit of the pre-trained Stable Diffusion Model to the Chinese landscape painting reached 26.78, and this serves as a benchmark. Then look at the model output after DreamBooth fine-tuning and LoRA fine-tuning, it is obvious that the FID of both is less than 26.78, which represents that the fine-tuned model is a better fit to the downstream domain of Chinese landscape painting. In the comparison of the two fine-tuned models, the output of the DreamBooth fine-tuned model has a lower FID, i.e., it fits the Chinese landscape painting dataset better.

Finally, in the inference stage, we asked 6 experts in Chinese landscape painting to rate the results of the 3 models (pre-trained Stable Diffuison Model, DreamBooth fine-tuned, and LoRA fine-tuned). Note that the input prompt (whether positive or negative) is the same for all three models' inferences, except that DreamBooth needs to add the unique identifier [V].

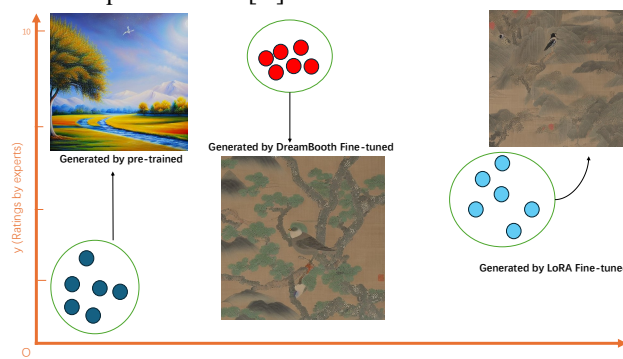

Figure 5. Expert ratings of images generated

Firstly, the generated image Prompt contains not only 'Chinese Landscape Paintings' but also 'A bird standing in a tree', which was chosen not only to observe how well the model understands Chinese Landscape Paintings, but also to consider the diversity, i.e., to retain the generic diversity of the original model. Pictures in Figure 5 were generated by the prompt 'A Chinese landscape painting with a bird standing on a tree'. And the whole figure represents the manual scoring of the images generated by the three models by six experts, and then the y-axis interval, is 0-10. The scoring results are then clustered into three cluster centres. Among the three models, the results generated by DreamBooth are more preferred by the experts, as its authenticity, diversity and fit with Chinese Landscape Paintings gave it this score. We also do manual scoring on other styles of landscape painting datasets, however, the result obtained is that the performance of the three models is similar.

## 5. CONCLUSION

In this paper, we investigate the performance of LoRA and DreamBooth, the current mainstream efficient fine-tuning methods, in generating Chinese Landscape Paintings using Stable Diffusion. Firstly, in the introduction section, we outline some of the Text-to-Image models and fine-tuning approaches on the landscape painting domain. Subsequently, in the Related Work section, we introduce Stable Diffusion, LoRA and DreamBooth and use them as comparative models in the Experimental section. In the methodology section, we present the essential principles of the Diffusion Model and Parameter-Efficient Fine-Tuning. In the experimental part, we compare the 3 models with FID evaluation metric and experts rating, which conclude that the Stable Diffusion model based on DreamBooth fine-tuning (SDDF) has a great potential on Chinese Landscape Paintings. However, the model has not been effectively represented in the performance of landscape paintings in other countries or style. So, in future work, we will continue to collect more datasets of this field to strengthen the validation and study the performance of DreamBooth SDM on landscape paintings in other countries.